\documentclass{poster23}
\usepackage{babel}%
\usepackage{amsmath}
\usepackage{pifont,amssymb}
\usepackage{graphicx}
\usepackage{wrapfig}

\begin{document}

%
\title{Influence of Depth Camera Noise Models on Respiration Estimation}
%
\headtitle{M. ROHR, S. DILL, INFLUENCE OF DEPTH CAMERA NOISE MODELS ON RESPIRATION ESTIMATION}

%
\author{Maurice ROHR\affiliationmark{1}, Sebastian DILL\affiliationmark{1}}
%
\affiliation{%
\affiliationmark{1}KIS*MED -- AI Systems in Medicine, Technische Universität Darmstadt, Merckstraße 25, Darmstadt, Germany }
  \email{rohr@kismed.tu-darmstadt.de, dill@kismed.tu-darmstadt.de}

\maketitle


\begin{abstract}
Depth cameras are an interesting modality for capturing vital signs such as respiratory rate. Plenty approaches exist to extract vital signs in a controlled setting, but in order to apply them more flexibly for example in multi-camera settings, a simulated environment is needed to generate enough data for training and testing of new algorithms. We show first results of a 3D-rendering simulation pipeline that focuses on different noise models in order to generate realistic, depth-camera based respiratory signals using both synthetic and real respiratory signals as a baseline.
While most noise can be accurately modelled as Gaussian in this context, we can show that as soon as the available image resolution is too low, the differences between different noise models surface.
\end{abstract}



\section{Introduction}

Camera-based sensing of vital signs is seeing an increase in popularity, as implied by the rising number of publications\footnote{https://pubmed.ncbi.nlm.nih.gov/?term=camera-based+vital+signs}. Possible reasons for that include the fact that contact-less estimation reduces the risk of transmitting diseases such as COVID~\cite{lewnard2020} and the possibility to exploit multiple sources of  biosignals such as skin perfusion, ballistocardiographic effects, thoracic/abdominal motion, body temperature and affordability of image sensors [2]. One common type of camera is depth cameras that estimate the distance of points in  a scene. They are especially useful in public screening settings as they provide inbuilt anonymisation (if raw RGB-camera streams are not recorded), while still enabling tracking of individuals for short time frames~\cite{imano2020,zhenpeng2014}. This is true for time-of-flight (ToF), stereo and structured light technologies. 
Even though new approaches to employing depth cameras are plenty, most studies use as few as 10 participants for training and evaluating their algorithms \cite{hoogi2019}. If many camera-configurations, and sensor fusion are to be tested, this is far from enough. At the same time, larger studies are expensive or might not even be feasible. This is where simulation can fill a gap. While simulating depth images is straight forward, the noise properties are highly dependent on the used device and even the environment~\cite{haider2022}. We propose a framework for simulating realistic sensor noise on depth cameras to enable the identification of respiratory patterns from depth camera information alone in difficult settings. We simulate realistic thoracic respiration on a human 3D model and show the influence of an array of depth camera specific noise sources on the raw signal. We then analyse the noise sources with respect to signal-to-noise ratio (SNR), since recent research emphasized the relevance of SNR to biosignal estimation~\cite{zaunseder2022}.

\section{Methods}

\begin{figure}[ht!]
\begin{center}
\resizebox{70mm}{!}{\includegraphics{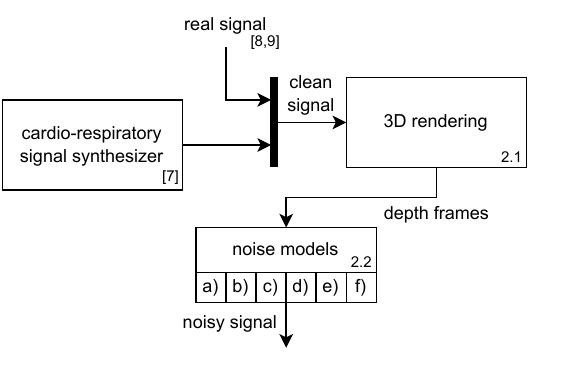}}
\caption{Overview of rendering and post-processing pipeline
} 
\label{fig:overview}
\end{center}
\end{figure}

Our method is composed of two steps (Fig. \ref{fig:overview}). We take both real and simulated respiratory signals which are proportional to the lung volume, and use them to modulate the thoracic movement. In the second step, we add artificial noise based on six different noise models, denoted by letters a-f. Finally, we extract the respiratory signal from a region-of-interest (RoI) from the chest and do a noise analysis.

\subsection{3D Rendering}
We used Blender\footnote{http://www.blender.org}, an open-source 3D-modelling and animation software, to simulate realistic human breathing. We took a publicly available 3D-model of a human male and modulated the chest and shoulder movement with (1) a synthetic respiratory signal generated with the framework proposed in~\cite{hoogi2017} and (2) a real respiratory signal from the fantasia database~\cite{iyengar1996,goldberger2000}. We employed linear resampling to adjust the sampling rate of the signal to the frame rate of the video. We also used a 4th order low-pass butterworth filter with a cut-off frequency of 1 Hz to smooth signal noise in the case of real signals.

\begin{figure}[ht!]
\begin{center}
\resizebox{45mm}{!}{\includegraphics{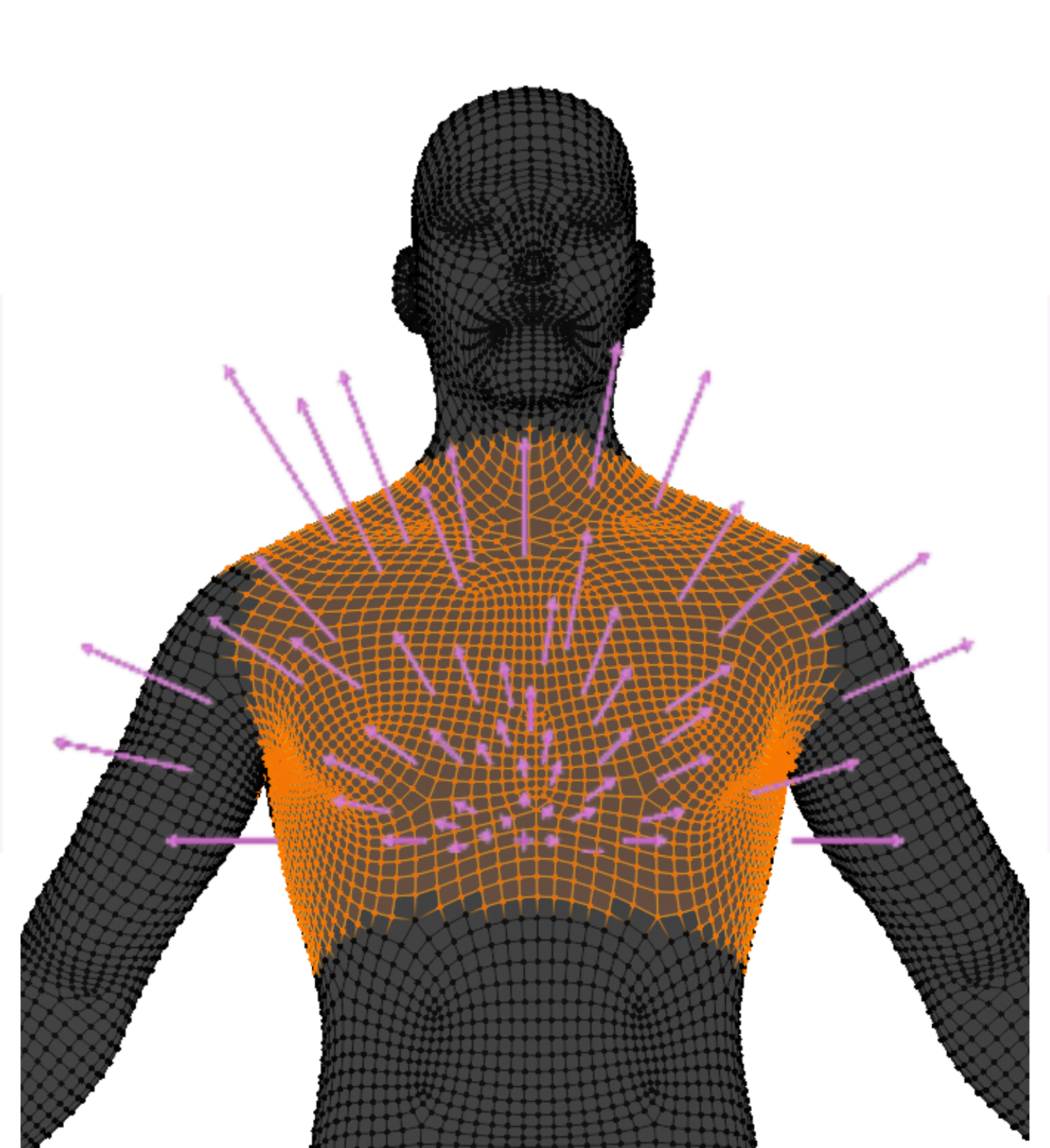}}
\caption{Body part mostly affected by thoracic respiration.
} 
\label{fig:chest_selection}
\end{center}
\end{figure}

Tsoli et al.~\cite{tsoli2014} showed that respiratory movement can be divided into abdominal and thoracic respiration simply using 3D reconstructions of camera footage of breathing humans and principal component analysis. We focused on thoracic respiration and chose the vertices included in that respiratory movement according to the vertices shown as relevant in the principal-component-analyis in~\cite{tsoli2014}. The vertices are also depicted in Fig. \ref{fig:chest_selection} with implied motion. We computed the center of gravity for these vertices and applied the motion in the general direction of a vector pointing up/forward ($[0,\frac{2}{3},\frac{1}{3}]$, normal to sternum)  similar to the chest breathing type in the respective paper, weighted with $w=(1-d/d_\text{max})^2$, where $d$ is the distance to the center of gravity of the relevant vertices and $d_\text{max}$ the maximum distance of the considered vertices.  
The normalized simulated or real signals were multiplied with a factor to adjust for breathing amplitude and multiplied with the motion vector. We generated 900 depth camera frames with a sampling rate of 30~Hz for both signals.

\subsection{Noise Models}
Based on~\cite{haider2022,mallick2014} we identified 5 important sources of sensor noise for depth cameras, which will be laid out in detail in the following. In general, there are 3 types of depth cameras: stereo-vision, time-of-flight, and structured light. Most unknown noise sources can be modelled as Gaussian because they are made up of multiple independent effects. That is why as a first model (a) we add Gaussian noise to the rendered depth images.
Given the pinhole camera model, the sensitivity of the estimated depth $Z$ can be expressed as $\sigma_Z = (m/f_b)Z^2 \sigma_p$ where $\sigma_p$ is the standard deviation of the normalized disparity according to~\cite{mallick2014}. Thus, the axial quantization noise (b) of a depth camera depends quadratically on the distance to the measured object. Based on empirical results, the noise distribution can be described as 
$\epsilon = \eta((Z-d_\text{offset})d_\text{level})^2$,
where $\eta$ is a normally distributed random variable.

\begin{figure}[ht!]
\begin{center}
\resizebox{70mm}{!}{\includegraphics{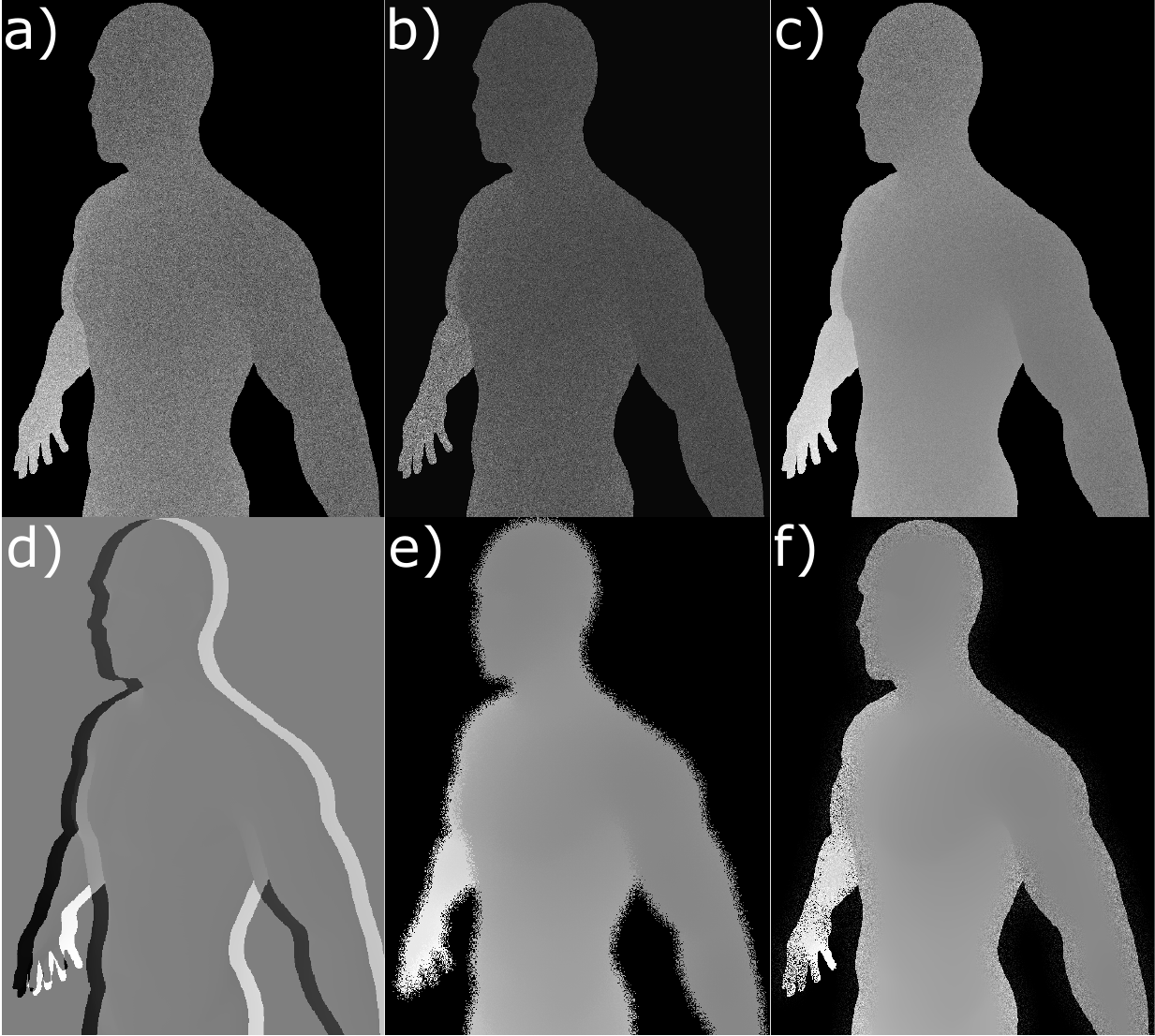}}
\caption{Visualization of applied depth camera noise types. Noise is exaggerated for visualization purposes. a) Gaussian noise, b) axial noise, c) radial noise, d) motion noise (difference image), e) edge permutation noise, f) edge Gaussian noise.
} 
\label{fig:noise_vis}
\end{center}
\end{figure}

Due to lens distortion effects, the measurement error also increases with the distance to the camera sensor which, in accordance with~\cite{haider2022}, we will call radial noise (c). We model radial noise as a linear faded circular mask between 0 (image center) and 1 (image corner) that adds weighted Gaussian noise to the depth image.
Motion blur has large effects on depth images~\cite{haider2022} and small RoI tracking errors during signal processing manifest themselves in a similar way. Thus, we apply random translation (d) to adjacent frames. 
A major noise source in time-of-flight and stereo depth images is small motions around strong depth edges such as object borders where triangulation might fail or the reflections are off by a few millimeters. We model this effect as a strong Gaussian noise (f) at object borders with the environment and what we call permutation noise (e), where beams originating from the correct object hit the image sensor at the wrong pixel location. It also implicitly models occlusion effects that usually happen at object borders, where the depth algorithms use interpolation to fill the resulting holes.

The process used to generate edge noise is simple: First we detect the edge of the respective object using as sobel filter and apply Gaussian smoothing with $\sigma_g$ to define the area of effect (AoE) of the noise. Then, we either directly apply strong Gaussian noise or resample pixel intensities uniformly from their surrounding intensities within a defined radius $r_p$ in the AoE to generate the permutation noise. The resulting noise, exaggerated for visualization purposes, is shown in Fig.\ref{fig:noise_vis}.

We analyse the signal inside a RoI on the chest of 280x206 pixels,which we also resampled to scales of 0.2 and 0.05 using nearest neighbour method.

\begin{figure}[ht!]
\begin{center}
\resizebox{50mm}{!}{\includegraphics[trim={0 3cm 0 2cm},clip]{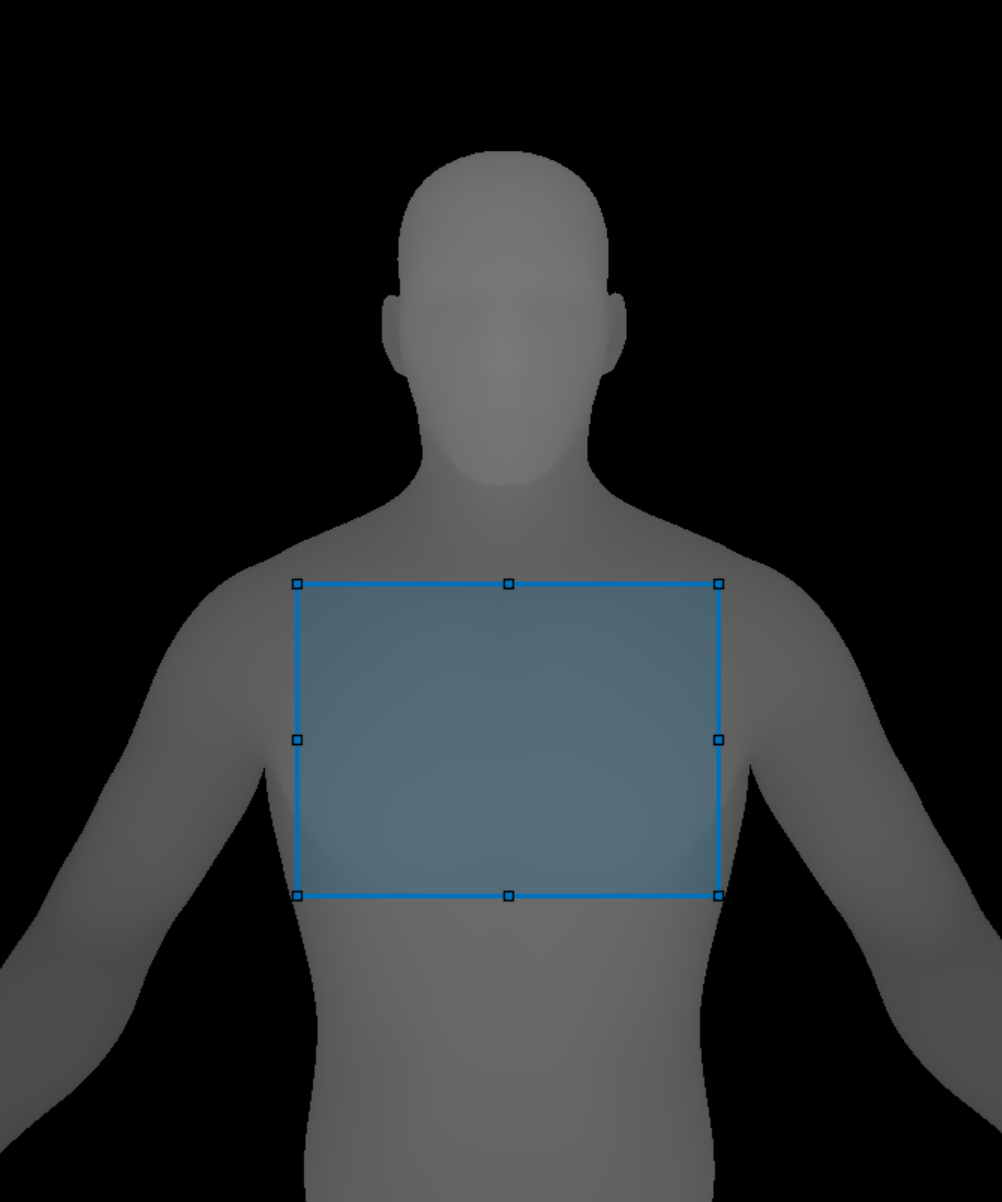}}
\caption{Select a RoI of the chest to extract signal.
} 
\label{fig:roi}
\end{center}
\end{figure}

\section{Results}

\begin{figure}[ht!]
\begin{center}
\resizebox{75mm}{!}{\includegraphics{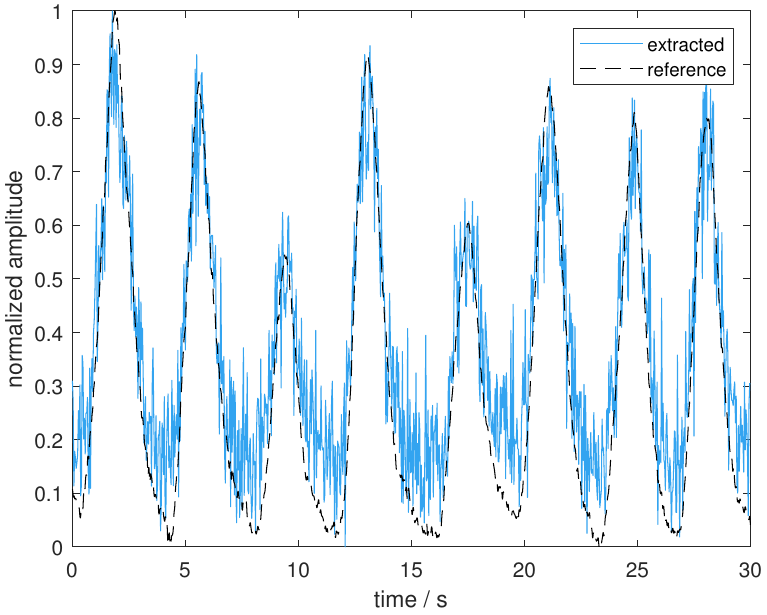}}
\caption{Fantasia reference signal and extracted noisy signal ($\sigma=0.067, \text{scale}=0.2, \text{SNR}=6.7$)
} 
\label{fig:signal_ex}
\end{center}
\end{figure}

\begin{figure}[ht!]
\begin{center}
\resizebox{80mm}{!}{\includegraphics{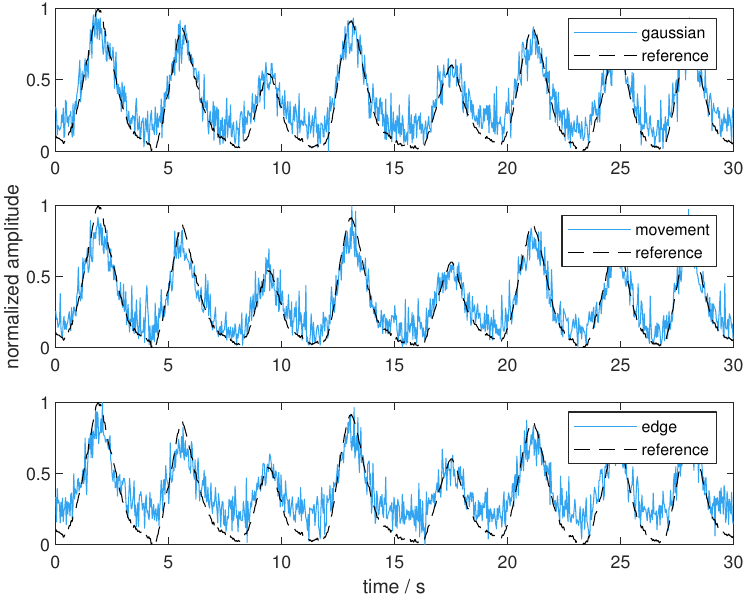}}
\caption{Fantasia reference signal and different types of noise for $\text{SNR}\approx7$
} 
\label{fig:signal_compare}
\end{center}
\end{figure}

There are no major differences between the effects of noise on the synthetic and real respiratory signals. In Fig.~\ref{fig:signal_ex} an example from the fantasia database is shown as reference to the noisy signal. For approximately same SNR and scale, the extracted noisy signals are qualitatively similar, independent of the presented noise models (Fig.~\ref{fig:signal_compare}).

\begin{figure}[ht!]
\begin{center}
\resizebox{85mm}{!}{\includegraphics{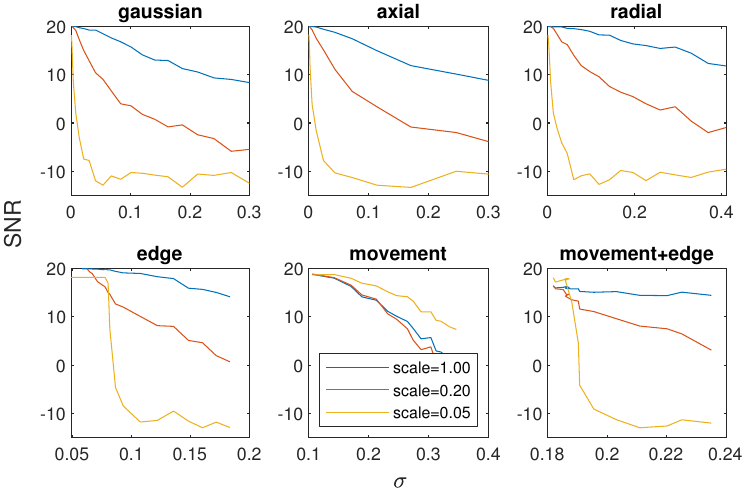}}
\caption{Influence of noise types on signal-to-noise ratio based on their standard deviation in the image on different image scales.
} 
\label{fig:noise_snr}
\end{center}
\end{figure}

We compute the SNR as the energy around the respiration rate and the first harmonic divided by the total energy of the signal.
Fig.~\ref{fig:noise_snr} shows the SNR-values compared to the standard deviation of the noise over the whole frame. The reason why we are comparing the SNR in relation to an artificial standard deviation $\sigma$, instead of the noise model parameters is that it leads to the same qualitative results, but complicates visualization. Axial noise and general Gaussian noise share a similar relation between SNR and $\sigma$ since the chest is approximately a flat surface parallel to the image plane, which reduces the effect of the quadratic dependency on the distance. Similarly, the radial noise shows the same trend. Especially for low resolution images, edge-noise deteriorates abruptly but has little effect for low margins ($\sigma_g$). Movement noise, as modelled here, is little affected by the scale and shows linear behaviour. The total effect of movement and edge noise for a fixed movement noise is additive. Additionally, movement decreases the margin where edge noise leads to high SNR.

\section{Discussion}

The observed similarity in effects of the different noise models can be explained by the central limit theorem. Since we apply approximately independent identically distributed noise inside a RoI that we then average, all noise models can be reparameterized as Gaussian.  This holds true as long as RoI placement is stable. For low resolutions, instable RoI tracking might lead to quick deterioration of the extracted signal. There are still a few things left to analyse: For rendering, we focused on the frontal view on the chest for analysis, which as we have shown, leads to Gaussian, axial and radial noise to behave very similarly. If a front view is not available, this might change heavily with increased angle and non-centered camera placement. The noise levels shown here are drawn from the realistic range, nevertheless measurements with the Kinect Sensor in~\cite{nguyen2012} show standard deviation of 0.04 for the axial noise, which is in the high SNR range in our experiment.

\section{Conclusion and Outlook}

We showcased our pipeline for 3d-rendering and synthesizing noisy depth videos for respiration estimation. We can generate realistic noisy respiratory signals for the evaluation of the respective algorithms. We can clearly see that in the end, Gaussian noise covers most of the effects and that sufficient resolution in the thorax area allows us to extract clean signals. If no averaging is used or the RoI is sufficiently small (e.g. by only using a single pixel), the noise model makes a difference, but further research is needed. We work on improving the noise models by directly including the sensor technologies in the 3D rendering engine such that also occlusion and similar effects can be directly captured and focus on small RoI applications, and then verifying the simulated results with recordings from different depth cameras. 

\section*{Acknowledgements}
The research described in the paper was supervised by Prof. C. Hoog Antink, TU Darmstadt.





\setlength{\intextsep}{0pt}%
\setlength{\columnsep}{2pt}%
\begin{authorcv}{}

\begin{wrapfigure}{l}{0.15\textwidth}\centering
\includegraphics[trim={2.5cm 0 2.5cm 0},clip,width=0.15\textwidth]{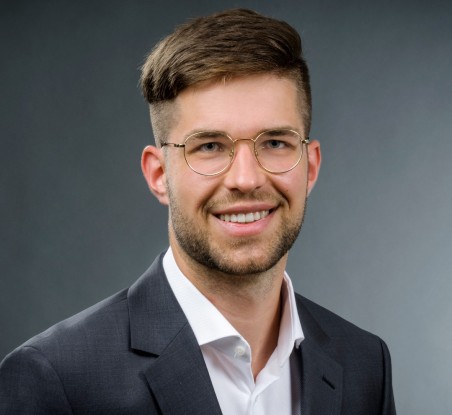}
\end{wrapfigure} 
\textbf{Maurice ROHR} was born in Mannheim, Germany, in 1995, finished his Abitur in 2014 and then began his studies in Darmstadt. In 2020 he finished his master degree in electrical engineering at the Technical University Darmstadt where he is currently working towards a Ph.D. degree in electrical engineering at the AI Systems in Medicine group. His research interests include medical sensor fusion, imaging technologies and simulation.

\begin{wrapfigure}{l}{0.15\textwidth}\centering
\includegraphics[trim={0.5cm 1cm 0 0.5cm},clip,width=0.15\textwidth]{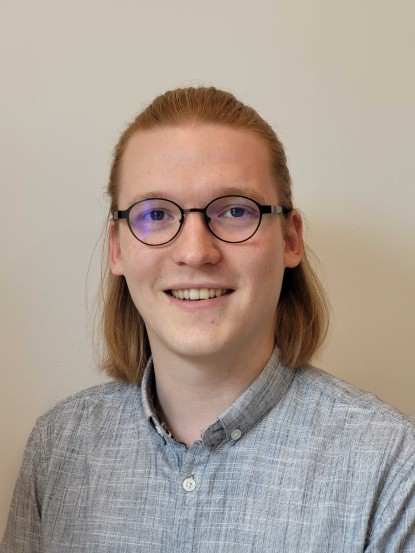}
\end{wrapfigure} 
\textbf{Sebastian DILL} was born in München, Germany, in 1996, finished his Abitur in 2014 and then began his studies in Darmstadt. In 2021 he finished his master degree in electrical engineering at the Technical University Darmstadt where he is currently working towards a Ph.D. degree in electrical engineering at the AI Systems in Medicine group. His research interests include non-obtrusive pose estimation, movement analysis and physical therapy.


\end{authorcv}

\end{document}